\documentclass[letterpaper]{article} % DO NOT CHANGE THIS
\usepackage{aaai2026}  % DO NOT CHANGE THIS
\usepackage{times}  % DO NOT CHANGE THIS
\usepackage{helvet}  % DO NOT CHANGE THIS
\usepackage{courier}  % DO NOT CHANGE THIS
\usepackage[hyphens]{url}  % DO NOT CHANGE THIS
\usepackage{graphicx} % DO NOT CHANGE THIS
\usepackage{natbib}  % DO NOT CHANGE THIS AND DO NOT ADD ANY OPTIONS TO IT
\usepackage{caption} % DO NOT CHANGE THIS AND DO NOT ADD ANY OPTIONS TO IT
\usepackage{algorithm}
\usepackage{algorithmic}
\usepackage{newfloat}
\usepackage{listings}
\usepackage{booktabs} % For professional looking tables
\usepackage{amsmath}  % For math if needed

\DeclareCaptionStyle{ruled}{labelfont=normalfont,labelsep=colon,strut=off} % DO NOT CHANGE THIS
\floatstyle{ruled}
\newfloat{listing}{tb}{lst}{}
\floatname{listing}{Listing}

\title{Comparative Analysis of Deep Learning Strategies for Hypertensive Retinopathy Detection from Fundus Images: From Scratch and Pre-trained Models}
\author {
    Yanqiao Zhu
}
\affiliations{
    Shanghai Jiao Tong University\\ % Or your specific affiliation
    1850432206@sjtu.edu.cn
}

\begin{document}

\maketitle

\begin{abstract}
This paper presents a comparative analysis of deep learning strategies for detecting hypertensive retinopathy from fundus images, a central task in the HRDC challenge~\cite{qian2025hrdc}. We investigate three distinct approaches: a custom CNN, a suite of pre-trained transformer-based models, and an AutoML solution. Our findings reveal a stark, architecture-dependent response to data augmentation. Augmentation significantly boosts the performance of pure Vision Transformers (ViTs), which we hypothesize is due to their weaker inductive biases, forcing them to learn robust spatial and structural features. Conversely, the same augmentation strategy degrades the performance of hybrid ViT-CNN models, whose stronger, pre-existing biases from the CNN component may be "confused" by the transformations. We show that smaller patch sizes (ViT-B/8) excel on augmented data, enhancing fine-grained detail capture. Furthermore, we demonstrate that a powerful self-supervised model like DINOv2 fails on the original, limited dataset but is "rescued" by augmentation, highlighting the critical need for data diversity to unlock its potential. Preliminary tests with a ViT-Large model show poor performance, underscoring the risk of using overly-capacitive models on specialized, smaller datasets. This work provides critical insights into the interplay between model architecture, data augmentation, and dataset size for medical image classification.
\end{abstract}

\begin{links}
    \link{Code}{https://github.com/byrTony-Frankzyq/HRDC_AutoMM}
    \link{Datasets}{https://codalab.lisn.upsaclay.fr/competitions/11877} % HRDC Challenge Link
\end{links}

\section{Introduction}

The field of computer vision has been revolutionized by deep learning, moving from hand-crafted features to end-to-end learning with models like AlexNet~\cite{krizhevsky2012imagenet}, ResNet~\cite{he2016deep}, and more recently, Vision Transformers (ViT)~\cite{dosovitskiy2020image}. ViTs, which apply the self-attention mechanism to image patches, have demonstrated state-of-the-art performance, particularly when pre-trained on large-scale datasets. This progress has profoundly impacted medical imaging, enabling automated systems for tasks like diabetic retinopathy detection~\cite{gulshan2016development} and skin cancer diagnosis~\cite{esteva2017dermatologist}.

The HRDC challenge~\cite{qian2025hrdc} addresses the critical task of detecting hypertension from fundus images. This paper explores a spectrum of deep learning strategies for this problem, from a simple custom network to advanced pre-trained models and AutoML. Our goal is to dissect the problem's complexity and understand the nuanced interactions between model architecture, pre-training, and data augmentation—a factor often treated as a universally beneficial pre-processing step.

The main contributions of this work are threefold:
\begin{itemize}
    \item We establish a critical baseline by showing that a simple custom CNN can achieve 100\% training accuracy, highlighting the model's capacity to overfit and emphasizing the need for robust generalization evaluation beyond training metrics.
    \item We conduct a systematic empirical study on a series of ViT and hybrid ViT-CNN models, revealing a strong, architecture-dependent relationship with data augmentation.
    \item We provide novel insights into fine-tuning transformers for hypertensive retinopathy detection, specifically:
        \begin{itemize}
            \item Uncovering that data augmentation is a double-edged sword: it is highly beneficial for pure ViTs, which lack strong inductive biases, but can be detrimental to hybrid models whose pre-trained CNN components may be disrupted by the transformations.
            \item Revealing that smaller patch sizes in pure ViTs (e.g., ViT-B/8) coupled with augmentation yield superior performance, likely by enhancing the model's ability to capture fine-grained pathological details.
            \item Demonstrating that advanced models, such as self-supervised DINOv2 and large-scale ViTs, require careful handling. DINOv2 fails catastrophically on the original data but is rescued by augmentation, while a ViT-Large model performs poorly, highlighting the crucial trade-off between model capacity and data sufficiency.
        \end{itemize}
\end{itemize}

\section{Related Work}

Our research is situated at the intersection of several key areas in computer vision and medical image analysis: the evolution of deep learning architectures, the role of pre-training strategies, and the application of automated machine learning frameworks.

\subsection{Evolution of Architectures: From CNNs to Transformers}

\paragraph{Convolutional Neural Networks (CNNs)} For nearly a decade, Convolutional Neural Networks have been the dominant architecture for computer vision tasks. Their design, which includes convolutional layers, pooling layers, and a hierarchical structure, introduces strong inductive biases such as translation equivariance and locality. These biases make them exceptionally efficient at learning visual patterns from images. Landmark models like AlexNet, VGGNet, GoogleNet, and especially ResNet with its residual connections to combat vanishing gradients, have become foundational tools. In medical imaging, these architectures have been successfully applied to a vast range of problems, including disease classification from X-rays, segmentation of tumors in MRI scans, and retinopathy detection from fundus images~\cite{gulshan2016development}. Our custom FundusCNN is based on these established principles to provide a simple, interpretable baseline.

\paragraph{Vision Transformers (ViT)} The paradigm shifted with the introduction of the Vision Transformer (ViT)~\cite{dosovitskiy2020image}, which adapted the highly successful Transformer architecture from Natural Language Processing to image recognition. By treating an image as a sequence of flattened patches, ViTs dispense with the hard-coded inductive biases of CNNs in favor of a global self-attention mechanism. This flexibility allows them to learn more complex, long-range dependencies within an image. However, this comes at the cost of being more "data-hungry"; early ViTs required pre-training on massive datasets like ImageNet-21k or JFT-300M to outperform their CNN counterparts. The models we investigate, such as ViT-B/8 and ViT-B/32, are direct descendants of this architectural lineage.

\paragraph{Hybrid CNN-Transformer Models} To bridge the gap between these two paradigms, researchers have developed hybrid architectures that aim to combine the best of both worlds. These models typically use a CNN-based stem to efficiently extract low-level features and capture local spatial information, which are then fed into a Transformer body to model global relationships. The ViT-B-R50/16 model we evaluate is a prime example, using a ResNet-50 backbone. More recent architectures like CAFormer~\cite{roy2023contextconvitformer} further integrate convolutional operations and context-aware mechanisms into the transformer blocks themselves, creating a more tightly coupled hybrid design. Our work directly investigates how these different architectural philosophies react to the same fine-tuning task and data augmentation strategies.

\subsection{The Role of Pre-training and Fine-tuning}

\paragraph{Supervised Pre-training and Transfer Learning} The standard practice for applying large models to specialized tasks is transfer learning. A model is first pre-trained on a large-scale, general-domain dataset like ImageNet, where it learns a rich hierarchy of visual features. The model is then fine-tuned on a smaller, task-specific dataset, such as the HRDC fundus images. This approach is highly effective because it leverages the "world knowledge" encoded in the pre-trained weights, dramatically reducing the amount of task-specific data and training time required. The PyTorch Image Models (`timm`) library has been instrumental in democratizing access to a wide array of such pre-trained models.

\paragraph{Self-Supervised Learning (SSL)} A limitation of supervised pre-training is its reliance on vast quantities of human-annotated labels. Self-supervised learning offers a powerful alternative by learning feature representations directly from unlabeled data. Methods like DINOv2~\cite{oquab2023dinov2} use objectives such as feature consistency across different augmented views of an image to learn robust and semantically meaningful features. The resulting models are excellent feature extractors that often transfer better to downstream tasks than their supervised counterparts, especially when the target domain (e.g., medical imaging) differs significantly from the pre-training domain (e.g., natural images). The dramatic performance shift of our DINOv2-based model highlights both the potential and the specific data requirements of this paradigm.

\subsection{Automated Machine Learning (AutoML)}
The process of selecting the right model architecture, optimizer, learning rate schedule, and other hyperparameters is complex and time-consuming. Automated Machine Learning (AutoML) aims to automate this process. Frameworks like AutoGluon~\cite{autogluon} encapsulate best practices from machine learning competitions and research into an easy-to-use tool. For vision tasks, AutoGluon-Multimodal leverages libraries like `timm` to provide access to state-of-the-art models and applies robust default HPOs and training procedures (e.g., using AdamW optimizer, cosine learning rate schedulers, and gradual unfreezing). By using AutoMM as our experimental framework, we not only accelerate our research but also ensure that our fine-tuning process is based on a strong, reproducible, and highly optimized baseline.

\section{Methodology}
\label{sec:methodology}

\subsection{A Simple CNN for Baseline Assessment}
\label{sec:simple_cnn}

To establish a baseline and probe the HRDC challenge's inherent difficulty, we designed FundusCNN, an exceptionally simple CNN with approximately \textbf{0.44 million} parameters. The architecture consists of four sequential convolutional blocks followed by a classifier. Each block contains a 2D Convolutional layer, Batch Normalization, Leaky ReLU, and Max Pooling. A final classifier head uses a $1 \times 1$ convolution, a fully connected layer, and Dropout for binary classification. As detailed in Section~\ref{sec:results_analysis}, this minimalistic model was capable of perfectly overfitting the training set, underscoring that the core patterns are learnable even by a low-capacity model and that generalization is the primary challenge. Further details are in our released code.

\subsection{Pre-trained Models and AutoML}
\label{sec:advanced_models}

Beyond the baseline, we investigated more sophisticated approaches:
\begin{enumerate}
    \item \textbf{Pre-trained Transformers:} We fine-tuned several ViT-based architectures initialized with public weights:
        \begin{itemize}
            \item Pure ViTs: ViT-Base/32, ViT-Base/8~\cite{dosovitskiy2020image}, and ViT-Base/14 pre-trained with DINOv2~\cite{oquab2023dinov2}.
            \item Hybrid ViT-CNNs: ViT-Base-R50/16 (a ViT with a ResNet-50~\cite{he2016deep} stem) and CAFormer-B36~\cite{roy2023contextconvitformer}.
        \end{itemize}
    \item \textbf{AutoML:} We used AutoGluon-Multimodal~\cite{autogluon} for automated model selection and hyperparameter tuning, providing a strong automated benchmark.
\end{enumerate}

\subsection{Data Augmentation Strategies}
\label{sec:data_augmentation}

We created three distinct dataset versions to test the impact of augmentation:
\begin{enumerate}
    \item \textbf{Original Dataset (Orig):} The raw HRDC training data.
    \item \textbf{Flipped Dataset (Flip):} Original data plus a horizontal flip of each image.
    \item \textbf{Augmented Dataset (Aug.):} The Flipped dataset further augmented with random transformations on-the-fly, including \textbf{Random Rotation} ($\pm 15^\circ$) and \textbf{Color Jitter} (brightness and contrast factors up to $\pm 0.3$).
\end{enumerate}
This strategy was designed to systematically evaluate the effect of adding simple geometric invariances versus more complex, randomized augmentations.

\section{Experiments and Results}
\label{sec:experiments_results}

\subsection{Experimental Setup}
\label{sec:experimental_setup}

All experiments used the official HRDC challenge training and validation splits. Fine-tuning and AutoML experiments were run using AutoGluon-Multimodal's default image classification settings. FundusCNN was trained for 100 epochs with an Adam optimizer, a learning rate of 1e-3, and a batch size of 64. Performance was measured with Cohen's Kappa (Kappa), F1-Score (Positive Class), Specificity (Negative Class), and an unweighted Average Score of the three.
\subsection{Advanced Models via AutoGluon-Multimodal}
\label{sec:autogluon_setup}
To explore more sophisticated architectures, we utilized AutoGluon-Multimodal (AutoMM)~\cite{autogluon}, an open-source AutoML framework that automates the process of building deep learning models for various data types, including images.

\paragraph{Automated Pipeline.} 
AutoMM streamlines the experimental process by automatically handling model selection, hyperparameter optimization (HPO), and fine-tuning. 
We leveraged its `MultiModalPredictor` class, specifying the problem type as binary classification and directing it to optimize for `kappa`.

\paragraph{Experimental Control.} 
% For our experiments, we constrained AutoMM to train a specific model at a time (e.g., vit_base_patch8_224 or caformer_b36.timm_in22k_ft_in1k) rather than using its default ensembling feature. 

For our experiments, we constrained AutoMM to train a specific model at a time rather than using its default ensembling feature.
This allowed us to isolate the performance of each architecture. 
We ran two sets of experiments for each model: one on the Original dataset and one on the Augmented dataset.
\paragraph{Default Hyperparameters and Training Tricks.} AutoMM incorporates a set of battle-tested default hyperparameters and training strategies designed for robust performance. We used these defaults for all transformer-based experiments. A summary of the key settings is provided in Table~\ref{tab:automm_hparams}. These defaults include advanced techniques like the AdamW optimizer for better regularization, a cosine learning rate scheduler for smooth convergence, and label smoothing to prevent model overconfidence. While AutoMM also supports advanced augmentations like MixUp, our study focused on the controlled application of geometric and color-based transformations to understand their direct impact.

\begin{table}[htbp]
\centering
\caption{Key Default Hyperparameters in AutoGluon-Multimodal.}
\label{tab:automm_hparams}
\begin{tabular}{@{}ll@{}}
\toprule
Hyperparameter & Default Value / Strategy \\
\midrule
Optimizer & AdamW \\
Learning Rate & 2e-4 (for ViT models) \\
LR Scheduler & Cosine Decay \\
Weight Decay & 0.05 \\
Batch Size & 16 (effective) \\
Warmup Steps & 5\% of total steps \\
Label Smoothing & 0.1 \\
Model Fine-tuning & Gradual unfreezing (head first) \\
\bottomrule
\end{tabular}
\end{table}

\subsection{Results Analysis}
\label{sec:results_analysis}

\begin{figure*}[htbp]
    \centering
    \begin{minipage}{0.48\textwidth}
        \centering
        \includegraphics[width=\linewidth]{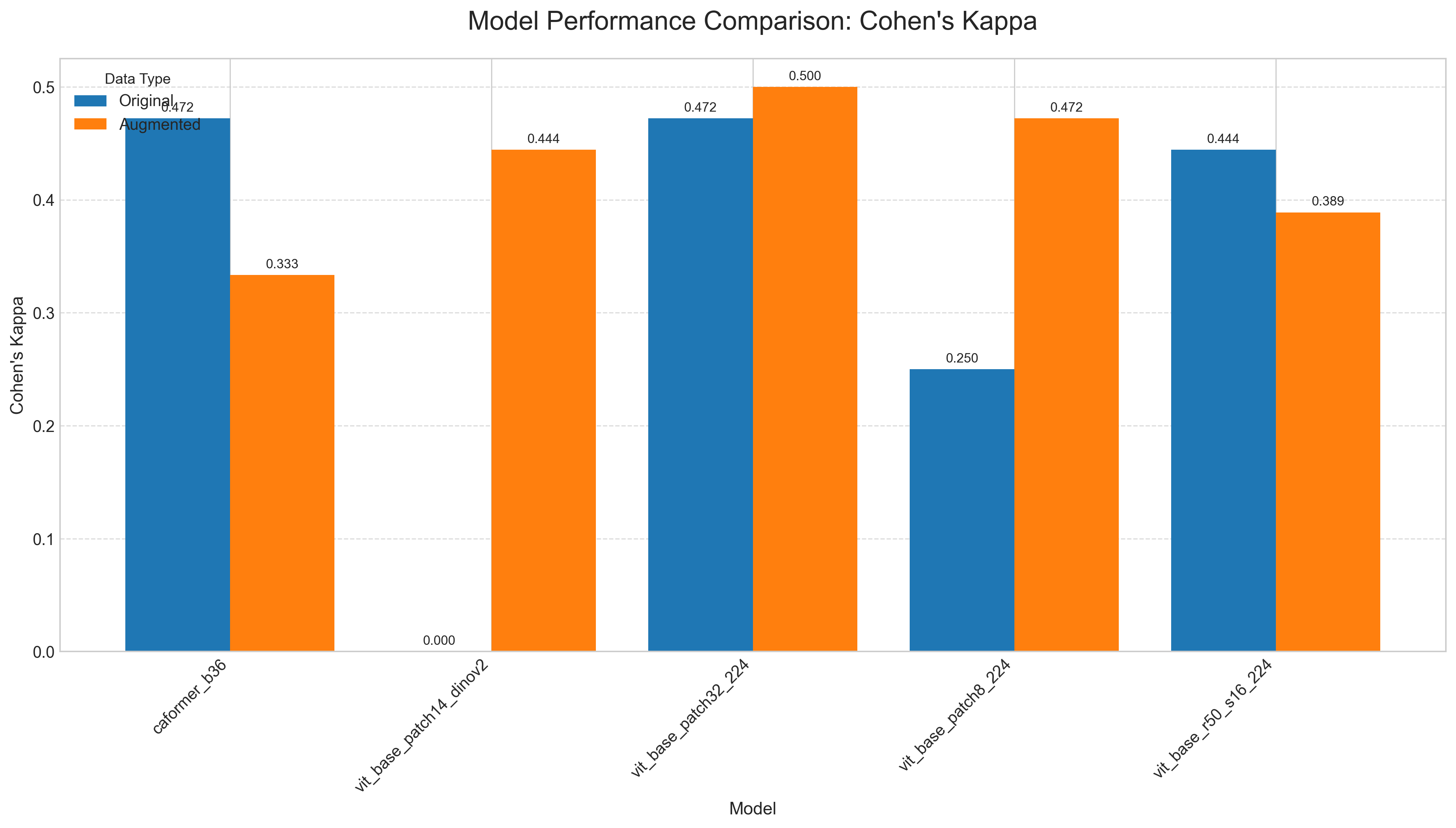}
        \caption{Cohen's Kappa comparison across models and datasets.}
        \label{fig:kappa_comparison}
    \end{minipage}\hfill
    \begin{minipage}{0.48\textwidth}
        \centering
        \includegraphics[width=\linewidth]{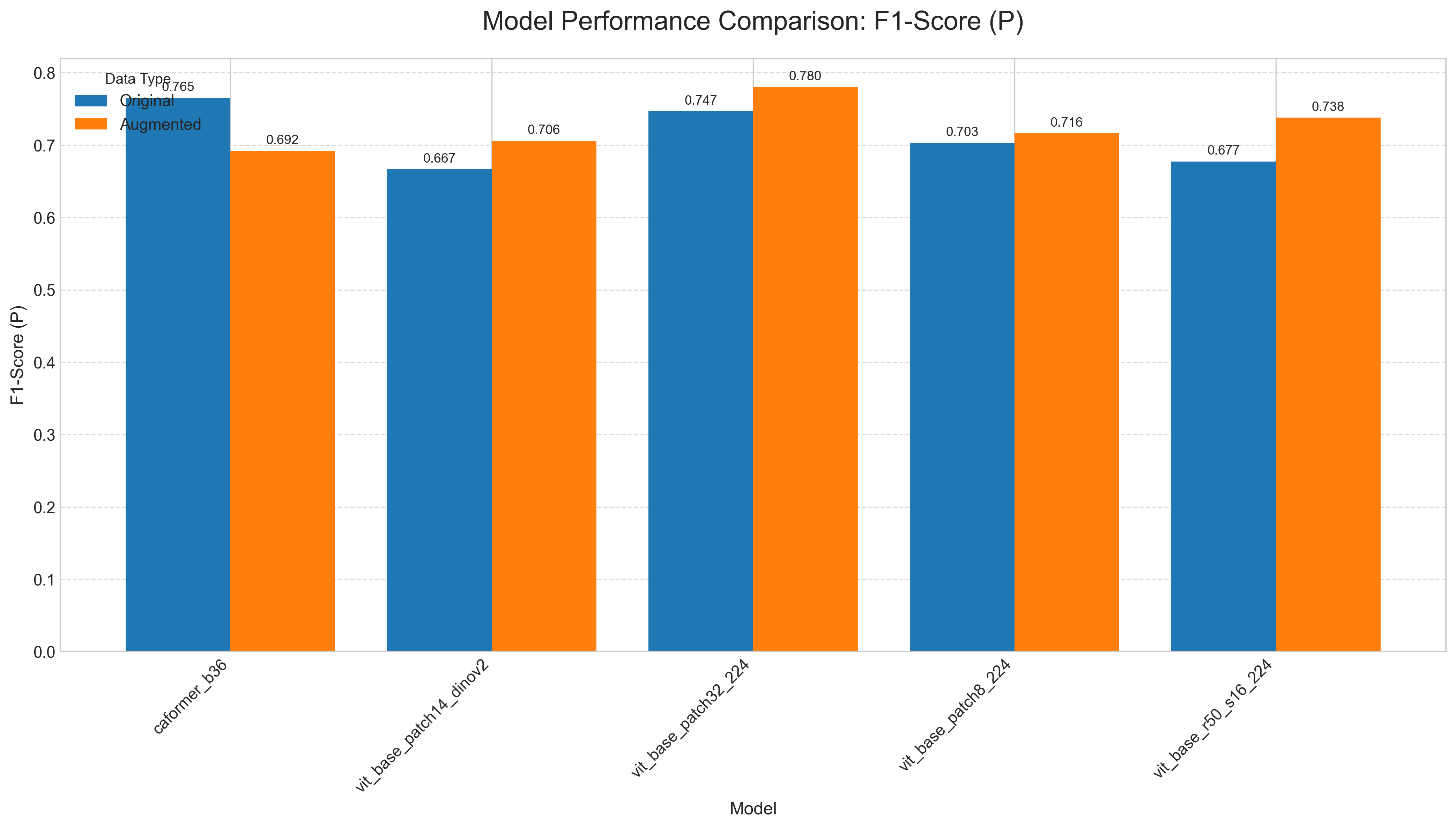}
        \caption{F1-Score (Positive Class) comparison.}
        \label{fig:f1_comparison}
    \end{minipage}
    \vspace{1em}
    \begin{minipage}{0.48\textwidth}
        \centering
        \includegraphics[width=\linewidth]{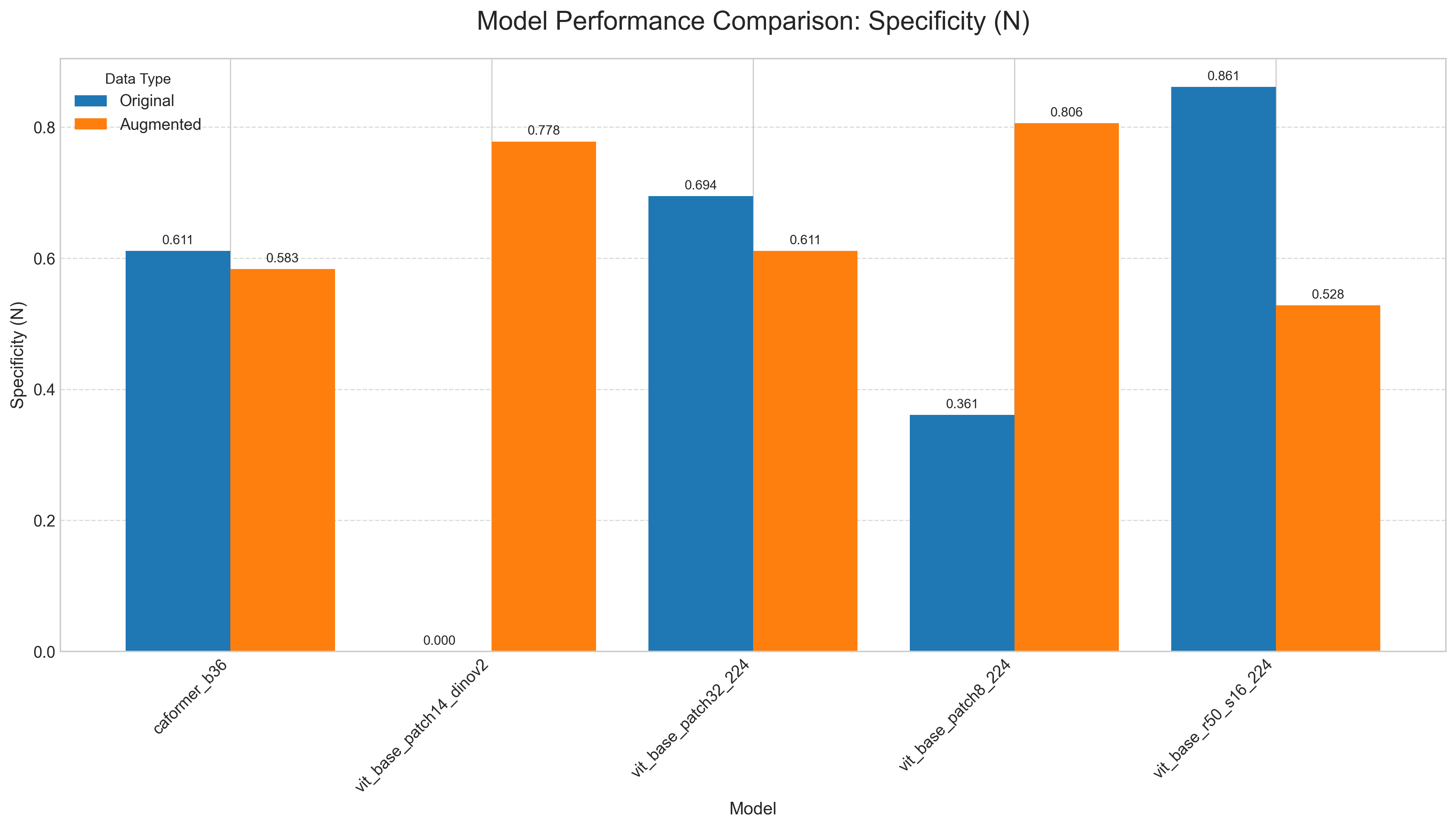}
        \caption{Specificity (Negative Class) comparison.}
        \label{fig:specificity_comparison}
    \end{minipage}\hfill
    \begin{minipage}{0.48\textwidth}
        \centering
        \includegraphics[width=\linewidth]{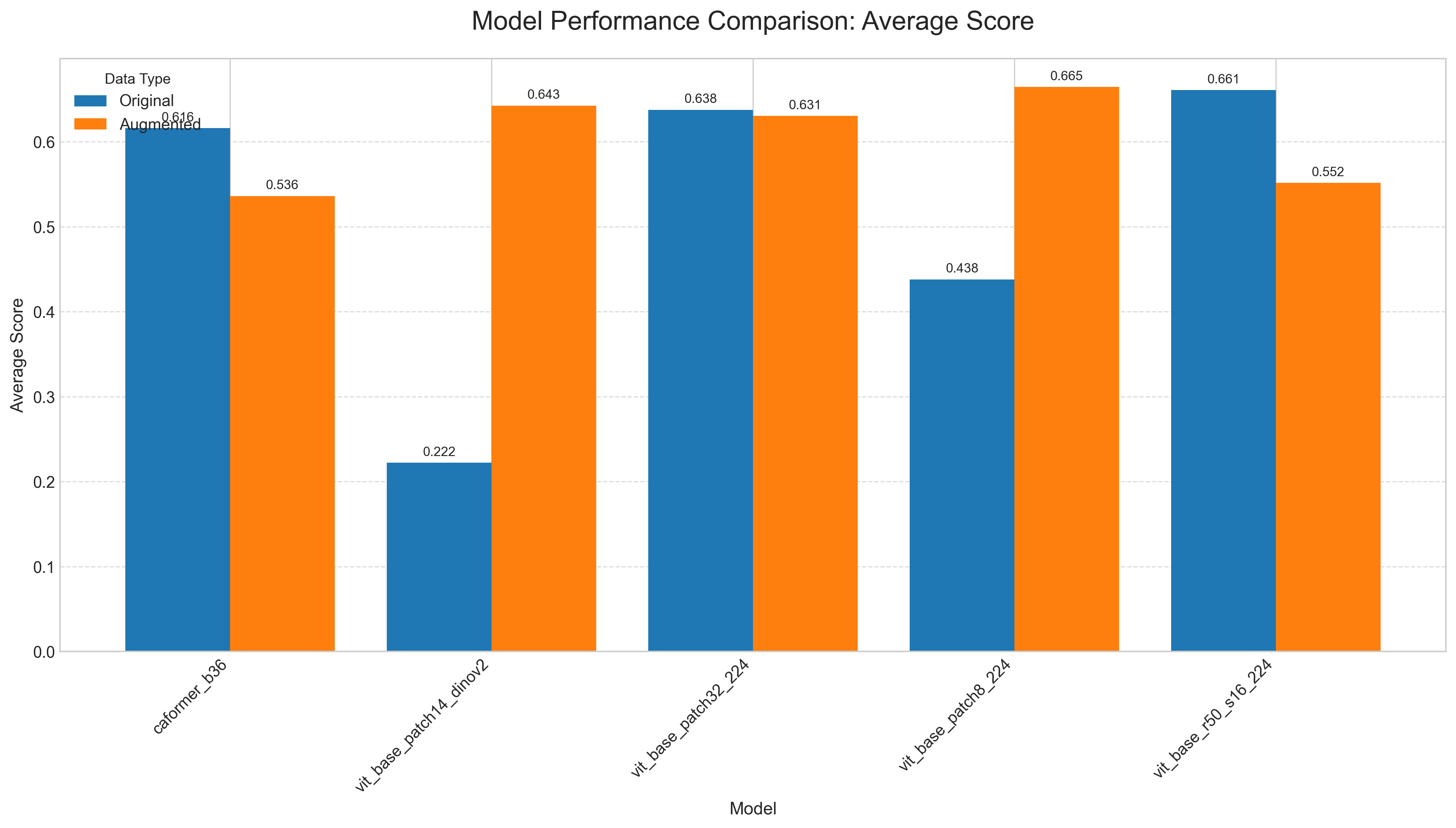}
        \caption{Average Score comparison.}
        \label{fig:avg_score_comparison}
    \end{minipage}
    \caption*{Figure~\ref{fig:kappa_comparison}-\ref{fig:avg_score_comparison}: Performance metrics for different models on original (blue/left bars) vs. augmented (orange/right bars) datasets. The divergent impact of augmentation is visually evident.}
    \label{fig:all_metrics_comparison}
\end{figure*}

Table~\ref{tab:results_comparison} and Figures~\ref{fig:kappa_comparison}-\ref{fig:avg_score_comparison} summarize model performance.

\begin{table*}[htbp]
\centering
\caption{Performance comparison of different models on original and augmented (Flipped + Random Aug) datasets for hypertension detection. (P) denotes the positive class (hypertension), (N) denotes the negative class. "Orig." refers to the original dataset, "Aug." to the Flipped + Random Aug dataset. Augmentation clearly helps pure ViTs while hurting hybrid models.} 
\label{tab:results_comparison}
\begin{tabular}{@{}llcccc@{}}
\toprule
Model & Data & Kappa & F1 (P) & Specificity (N) & Avg. Score \\
\midrule
CAFormer-B36 (Hybrid) & Orig. & 0.472 & 0.765 & 0.611 & 0.616 \\
CAFormer-B36 (Hybrid) & Aug. & 0.333 & 0.692 & 0.583 & 0.536 \\
\midrule
ViT-B-R50/16 (Hybrid) & Orig. & 0.444 & 0.677 & 0.861 & 0.661 \\
ViT-B-R50/16 (Hybrid) & Aug. & 0.389 & 0.738 & 0.528 & 0.552 \\
\midrule
ViT-B/8 (Pure) & Orig. & 0.250 & 0.703 & 0.361 & 0.438 \\
ViT-B/8 (Pure) & Aug. & \textbf{0.472} & 0.716 & \textbf{0.806} & \textbf{0.665} \\
\midrule
ViT-B/14-DINOv2 (Pure) & Orig. & 0.000 & 0.667 & 0.000 & 0.222 \\
ViT-B/14-DINOv2 (Pure) & Aug. & 0.444 & 0.706 & 0.778 & 0.643 \\
\midrule
ViT-B/32 (Pure) & Orig. & 0.472 & 0.747 & 0.694 & 0.638 \\
ViT-B/32 (Pure) & Aug. & 0.500 & 0.780 & 0.611 & 0.631 \\
\bottomrule
\end{tabular}
\end{table*}

\paragraph{FundusCNN Baseline.}
Our simple FundusCNN achieved 100\% accuracy on the training set, confirming it could memorize the data. This highlights that the primary difficulty of the HRDC task lies in generalization, not in learning the training distribution itself.

\paragraph{Divergent Impact of Data Augmentation.}
The results show a clear dichotomy.
\begin{itemize}
    \item \textbf{Pure ViT Architectures Benefit:} ViT-B/8 and ViT-B/14-DINOv2 showed dramatic improvements with augmentation. ViT-B/8's Kappa soared from 0.250 to 0.472. Most strikingly, ViT-B/14-DINOv2, which completely failed on the original data by predicting only the positive class (Kappa 0.000), was "rescued" by augmentation, achieving a respectable Kappa of 0.444.
    \item \textbf{Hybrid Architectures Suffer:} In stark contrast, both hybrid models saw performance drops. CAFormer-B36's Kappa fell from 0.472 to 0.333, and ViT-B-R50/16's Kappa fell from 0.444 to 0.389.
\end{itemize}

\paragraph{Architectural Granularity and Patch Size.}
With augmentation, the pure ViT with a smaller patch size, \textbf{ViT-B/8}, achieved the highest Average Score (0.665). Its performance gain was much larger than that of the larger-patch ViT-B/32. This suggests that smaller patches are more effective at capturing fine-grained pathological details in fundus images, provided the model has enough data (via augmentation) to learn the broader spatial context.

\paragraph{Challenges with Model Scaling.}
We performed preliminary experiments with a larger ViT-Large model (not shown in the table). It performed poorly across all metrics, significantly worse than the Base-sized models. This is a classic example of a model-data mismatch; on a relatively small dataset like HRDC, the larger model likely failed to train effectively or overfit severely, reinforcing the principle that bigger is not always better in specialized domains.

\section{Discussion}
\label{sec:discussion}

Our results, particularly the opposing effects of data augmentation, merit a deeper discussion. We propose the following interconnected hypotheses.

\paragraph{1. Inductive Bias and the Data Appetite of Pure ViTs.}
Pure ViTs lack the built-in translational equivariance and locality bias of CNNs. By design, they treat an image as a sequence of patches, making the task of modeling spatial relationships more challenging. We hypothesize that data augmentation acts as a powerful regularizer that is particularly crucial for these architectures. It forces the model to learn features that are invariant to transformations like flips, rotations, and color shifts, compelling it to focus on the underlying structure of the retinal vasculature rather than superficial texture or orientation. This explains why augmentation provided a substantial boost, especially for a model that starts with a weaker task-specific prior.

\paragraph{2. Feature "Confusion" in Hybrid Architectures.}
In contrast, hybrid models like ViT-B-R50/16 and CAFormer-B36 already possess strong inductive biases from their CNN components. These components are pre-trained on natural images where certain features and spatial relationships are consistently learned. We posit that our aggressive augmentation strategy (especially random rotations and jitter) may introduce transformations that conflict with the features the CNN part is optimized to extract. This "confuses" the model, disrupting the delicate interplay between the CNN feature extractor and the transformer body, ultimately leading to performance degradation.

\paragraph{3. Unlocking Self-Supervised Models: The DINOv2 Case.}
The behavior of ViT-B/14-DINOv2 is highly illustrative. Its complete failure on the original dataset suggests that while self-supervised pre-training yields powerful, general-purpose feature extractors, these features are not immediately plug-and-play. On a small, homogenous dataset, the model may latch onto spurious correlations. Augmentation provided the necessary data diversity to regularize the fine-tuning process, forcing the model to adapt its robust general features to the specific pathology of hypertensive retinopathy, thereby unlocking its potential.

\paragraph{Implications for Practice.}
These findings have direct practical implications. The choice of data augmentation strategy should not be generic but rather co-designed with the model architecture. For pure ViTs on medical images, augmentation appears not just beneficial but necessary. For hybrid models, a more conservative or targeted augmentation approach may be required. Furthermore, the success of the smaller-patch ViT-B/8 and the failure of ViT-Large serve as a strong reminder to prioritize architectural choices and model scale appropriate for the dataset size over simply using the largest available model.

\section{Conclusion}
\label{sec:conclusion}

This paper presented a comparative analysis of deep learning strategies for hypertensive retinopathy detection, yielding several key insights. We established that the HRDC challenge is primarily a test of generalization, as even a simple CNN can overfit the training data. Our central finding is the starkly different, architecture-dependent impact of data augmentation: it is essential for regularizing and improving pure ViT models but can be detrimental to hybrid ViT-CNN architectures. We attribute this to the interplay between the model's inherent inductive biases and the nature of the transformations. Furthermore, we demonstrated that smaller patch sizes can be advantageous for capturing fine-grained medical details and that powerful pre-trained models like DINOv2 and ViT-Large require careful handling and sufficient data diversity to be effective. These results underscore that a "one-size-fits-all" approach to model selection and data augmentation is suboptimal for medical imaging tasks and provide a clear path for future research in developing more robust and tailored solutions.

\bibliography{aaai2026} % Assuming your bib file is named aaai2026.bib

\end{document}